\documentclass[sigconf]{acmart}

\AtBeginDocument{%
  }

\copyrightyear{2025}
\acmYear{2025}
\setcopyright{acmlicensed}
\acmConference[MM '25] {Proceedings of the 33rd ACM International Conference on Multimedia}{October 27--31, 2025}{Dublin, Ireland.}
\acmBooktitle{Proceedings of the 33rd ACM International Conference on Multimedia (MM '25), October 27--31, 2025, Dublin, Ireland}
\acmISBN{979-8-4007-2035-2/2025/10}
\acmDOI{10.1145/3746027.3754504}

\usepackage{cite}
\usepackage{svg}
\usepackage{amsmath}
\usepackage{algorithmic}
\usepackage{textcomp}
\usepackage{xcolor}
\usepackage{color}
\usepackage{multirow}
\usepackage{graphicx}
\usepackage{balance}
\usepackage{subfigure}
\usepackage{array}
\usepackage{colortbl}
\usepackage{booktabs}
\usepackage{mathrsfs}
\usepackage{mfirstuc}
\usepackage{makecell}
\usepackage{algorithm} %format of the algorithm 
\usepackage{algorithmic} %format of the algorithm
\usepackage{hyperref}
\begin{document}

\title{Perspective from a Higher Dimension: Can 3D Geometric Priors Help Visual Floorplan Localization?}

\author{Bolei Chen}
\email{boleichen@csu.edu.cn}
\affiliation{%
  \institution{School of Computer Science and \\ Engineering, Central South University}
  \city{Changsha}
  \state{Hunan}
  \country{China}
}

\author{Jiaxu Kang}
\email{jxkang@csu.edu.cn}
\affiliation{%
  \institution{School of Computer Science and \\ Engineering, Central South University}
  \city{Changsha}
  \state{Hunan}
  \country{China}
}

\author{Haonan Yang}
\email{haonanyang@csu.edu.cn}
\affiliation{%
  \institution{School of Computer Science and \\ Engineering, Central South University}
  \city{Changsha}
  \state{Hunan}
  \country{China}
}

\author{Ping Zhong}
\authornotemark[1]
\email{ping.zhong@csu.edu.cn}
\affiliation{%
  \institution{School of Computer Science and \\ Engineering, Central South University}
  \city{Changsha}
  \state{Hunan}
  \country{China}
}

\author{Jianxin Wang}
\authornote{Corresponding authors.}
\email{jxwang@mail.csu.edu.cn}
\affiliation{%
  \institution{School of Computer Science and \\ Engineering, Central South University}
  \city{Changsha}
  \state{Hunan}
  \country{China}
}

% \renewcommand{\shortauthors}{Bolei Chen et al.}
\renewcommand{\shortauthors}{Bolei Chen, Jiaxu Kang, Haonan Yang, Ping Zhong, and Jianxin Wang}
\renewcommand{\shorttitle}{Can 3D Geometric Priors Help Visual Floorplan Localization?}

\begin{teaserfigure}
 \centering
 \includegraphics[scale=1.07]{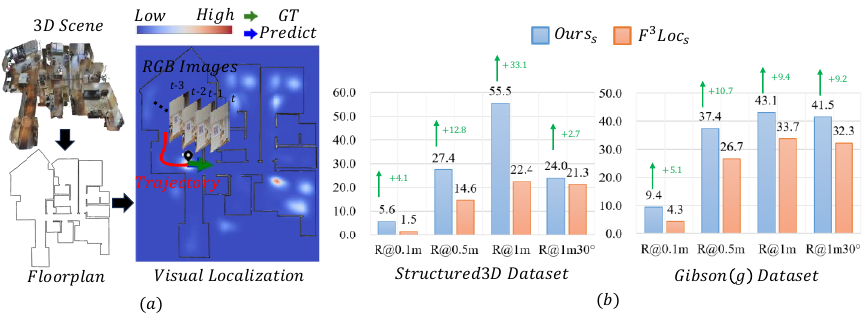}
 \vspace{-0.5cm}
 \caption{(a) An illustration of the visual Floorplan Localization (FLoc) task. (b) Our 3D geometric prior-enhanced visual FLoc method significantly improves the localization success and accuracy relative to the SoTA method $F^3Loc$ \citep{chen2024f3loc} on two challenging datasets (Structured3D \citep{zheng2020structured3d} and Gibson \citep{xia2018gibson}).}
 \Description{(a) An illustration of visual floorplan localization. (b) Our 3D geometric prior-enhanced visual floorplan localization method significantly improves the localization success and accuracy relative to the SoTA method $F^3Loc$ on two challenging datasets.}
 \label{fig1}
\end{teaserfigure}

\begin{abstract}
 Since a building's floorplans are easily accessible, consistent over time, and inherently robust to changes in visual appearance, self-localization within the floorplan has attracted researchers' interest. However, since floorplans are minimalist representations of a building's structure, modal and geometric differences between visual perceptions and floorplans pose challenges to this task. While existing methods cleverly utilize 2D geometric features and pose filters to achieve promising performance, they fail to address the localization errors caused by frequent visual changes and view occlusions due to variously shaped 3D objects. To tackle these issues, this paper views the 2D \textbf{F}loorplan \textbf{Loc}alization (FLoc) problem from a higher dimension by injecting 3D geometric priors into the visual FLoc algorithm. For the 3D geometric prior modeling, we first model geometrically aware view invariance using multi-view constraints, i.e., leveraging imaging geometric principles to provide matching constraints between multiple images that see the same points. Then, we further model the view-scene aligned geometric priors, enhancing the cross-modal geometry-color correspondences by associating the scene's surface reconstruction with the RGB frames of the sequence. Both 3D priors are modeled through self-supervised contrastive learning, thus no additional geometric or semantic annotations are required. These 3D priors summarized in extensive realistic scenes bridge the modal gap while improving localization success without increasing the computational burden on the FLoc algorithm. Sufficient comparative studies demonstrate that our method significantly outperforms state-of-the-art methods and substantially boosts the FLoc accuracy. 
 
 % All data and code will be released after the anonymous review.
  
\end{abstract}




%\received{20 February 2024}
%\received[revised]{12 March 2024}
%\received[accepted]{5 June 2024}

\maketitle

\vspace{-0.3cm}
\section{Introduction}

\textbf{F}loorplan \textbf{Loc}alization (FLoc) \citep{chen2024f3loc} is an interesting research topic that opens new possibilities for robotic navigation \citep{li2024flona} and AR/VR applications for mobile devices. Humans can visually localize themselves in 3D environments using simple 2D floorplans. However, visual localization algorithms rely heavily on pre-collected databases \citep{balntas2018relocnet, 2017NetVLAD} or complex 3D scene reconstructions \citep{liu2017efficient, sarlin2019coarse, sattler2016efficient}, which are expensive to build, store, and maintain. Since floorplan is lightweight, consistent over time, and inherently robust to changes in visual appearance, we aim to bridge the above-mentioned gap by investigating geometric prior-enhanced visual FLoc techniques. Such floorplans, which can be found in places such as shopping malls, office buildings, train stations, and apartments, encode rich and sufficient information to aid visual localization in unvisited scenes, as shown in Fig. \ref{fig1} (a). 

However, the geometric and modal differences between RGB images and floorplans pose challenges to visual FLoc tasks. On the one hand, floorplan modality often contains many repetitive structures, such as hallways and corners, because of its minimalist and compact form. These repetitive geometric structures can easily cause ambiguous or even incorrect localization. On the other hand, the visual modality is filled with category-rich, variously shaped, and intricately arranged 3D objects that are not reflected in the floorplans. The occlusion of the line of sight by 3D objects inevitably leads to localization biases. Objects with similar appearances and textures can easily cause mislocalization due to confusion of visual features, e.g., a bed and a sofa look visually similar. 

%介绍现有的方法在利用2D几何特征和基于滤波器的序列定位方面的努力，效果不佳
To address the above issues, existing work makes efforts in 2D geometric priors exploitation \citep{chu2015you, boniardi2019robot, howard2022lalaloc++, howard2021lalaloc, min2022laser} and Bayesian filtering-based sequential localization \citep{chu2015you, boniardi2019robot, ito2014w, karkus2018particle, chen2024f3loc}. 2D geometric features extracted from RGB images can be directly matched to floorplans to bridge the modal gap. Nevertheless, previous methods either assume that the camera and ceiling heights are known \citep{chu2015you, boniardi2019robot, howard2021lalaloc} or require panoramic images \citep{howard2021lalaloc, howard2022lalaloc++}, which inhibit the widespread usage of FLoc algorithms. While Laser \citep{min2022laser} designs a better observation model to gather and represent 2D features, its localization performance is greatly reduced when scene semantics are absent. Since image sequences \citep{sarlin2022lamar, sarlin2023orienternet} can somewhat eliminate the localization ambiguity caused by repetitive structures, other methods \citep{karkus2018particle, ito2014w, chen2024f3loc} optimize the posterior distribution of the current pose using Bayesian filtering. Among them, F$^3$Loc \citep{chen2024f3loc} is a classic FLoc framework consisting of a front-end 2D visual encoder and a back-end Bayesian filter, which achieves the \textbf{S}tate-\textbf{o}f-\textbf{T}he-\textbf{A}rt (SoTA) FLoc performance. However, F$^3$Loc focuses on the design of the back-end filter and neglects the effective representation of visual features, thus still achieving unsatisfactory FLoc performance.
%接着写在观测模型方面做出努力，以提取3D几何先验，同时讨论由物体导致的定位偏差，这是现有的方法没有讨论的

In this paper, by injecting 3D geometric priors into the observation model, our core idea is to provide FLoc with better feature bases by leveraging higher dimensional visual representations. Therefore, we propose to model two types of 3D geometric priors, including \textbf{G}eometry-\textbf{C}onstrained \textbf{V}iew \textbf{I}nvariance (GCVI) and \textbf{V}iew-\textbf{S}cene \textbf{A}ligned \textbf{G}eometric (VSAG) prior. Specifically, we first model GCVI by utilizing multi-view constraints, i.e., leveraging imaging geometric principles to provide matching constraints between multiple images that see the same points. Then, we further model the VSAG prior, enhancing the cross-modal geometry-color correspondences by associating the scene's surface reconstruction with the RGB frames of the sequence. Both geometric priors can be learned by contrastively pre-training on publicly available 3D scene datasets, such as the ScanNet \citep{dai2017scannet} RGB-D dataset. The observation model integrated with 3D geometric priors is fine-tuned end-to-end along with a Bayesian filter \citep{jonschkowski2016end} to fit the FLoc task. Our FLoc method is proposed based on the classic front- and back-end framework \citep{chen2024f3loc}, which localizes by finding the pose in the floorplan that has the most similar 2D rays as the prediction, as shown in Fig. \ref{fig2}.

Our 3D geometric priors benefit FLoc in the following aspects: \textit{\textbf{(1) More robust localization.}} Existing FLoc methods do not explicitly discuss and address localization biases caused by frequent visual changes and 3D object-induced view occlusions. In our work, GCVI is learned to enhance FLoc's robustness to visual changes by utilizing the content invariance in multiple views. The VSAG prior is designed to align views and scenes, making our FLoc method more sensitive to the geometric and visual appearance of 3D objects. \textit{\textbf{(2) More accurate localization.}} Although other scene prior modeling methods \citep{chen2020simple, du2021curious, hong2023learning, chen2024embodied} exist, their geometric constraints are loose, e.g., contrastive learning is performed only at the image or semantic level. In other words, their feature embeddings are image/point cloud-aligned rather than pixel/point-aligned. For FLoc tasks with high accuracy requirements, we constrain positive pairs for contrastive learning at the pixel/point level, e.g., less than 2 cm error for multiple observations of a point. Such a hard constraint makes our 3D priors more suitable for accurate FLoc. \textit{\textbf{(3) More efficient modal alignment.}} Unlike existing methods \citep{chu2015you, boniardi2019robot, howard2022lalaloc++, howard2021lalaloc, min2022laser} based on 2D geometric cues, our main insight is to associate visual features with 3D space using imaging geometric constraints and further project them as 2D rays to match the floorplan. This motivation for bridging the modal gap derives from human intuition for localization in a 3D scene. Please imagine that you wisely project the corner of the living room you see to the 3D outline (walls) and further down to the X-Y plane to align with the floorplan. \textit{\textbf{(4) No need for expensive labels and annotations.}} Our geometric prior modelings are achieved using self-supervised contrastive learning, thus requiring no expensive labels and annotations. The 3D geometric priors are injected into the observation model without increasing the computational burden on FLoc.

We conduct sufficient comparative studies between our method and strong baselines on the challenging Structured3D \citep{zheng2020structured3d} and Gibson \citep{xia2018gibson} datasets. Our method achieves SoTA visual FLoc performance and substantially outperforms the strong baseline, as shown in Fig. \ref{fig1} (b). In addition, ablation and qualitative studies demonstrate the superiority and potential of our 3D geometric priors in the FLoc task. 

\vspace{-0.3cm}
\section{Related Work}

\subsection{Visual Localization}
Traditional methods are achieved by explicitly \citep{sun2021loftr, yang2019sanet, zhou2021patch2pix, sarlin2021back} or implicitly \citep{balntas2018relocnet, ding2019camnet, kendall2015posenet} corresponding visual appearance with scene representations. Such scene representations are usually sparse/dense 3D reconstructions or databases composed of images with known camera poses. For example, image retrieval-based methods \citep{2017NetVLAD, balntas2018relocnet} find the most similar image from a database and use its pose as the localization. Besides, the camera pose can be recovered using 3D SfM models \citep{sarlin2019coarse, sattler2016efficient} from 2D-3D correspondences. Unlike these classical methods, data-driven models utilize neural networks to regress the 3D coordinates of the pixels in the query image \citep{valentin2015exploiting, brachmann2017dsac} or directly regress the 6D camera pose of the input image \citep{walch2017image, wu2017delving}. However, the appearance dependency limits the robustness of visual localization algorithms to scene changes, making them unable to handle purely geometric maps with missing appearances. In addition, these methods strongly rely on pre-built image databases and 3D scene models, which are expensive to store and maintain.

Recently, \textbf{Ne}ural \textbf{R}adiation \textbf{F}ield (NeRF)-based methods \citep{mildenhall2021nerf, kwon2023renderable, wang2024lookahead, liu2024integrating} open new technical routes to synthesize realistic images from scene-specific neural representations learned using back-propagation. The rendered images are matched with the query image for visual localization. These methods \citep{kwon2023renderable, wang2024lookahead, liu2024integrating} use NeRF to internalize visual features into a neural network to achieve an implicit scene representation, which serves as the feature base for visual localization. In this work, we internalize 3D geometric priors in the FLoc algorithm's observation model via self-supervised contrastive learning to improve the visual FLoc performance. Our 3D prior modeling does not rely on visual semantics and requires no maintenance of costly 3D scene reconstructions and databases.

\vspace{-0.4cm}
\subsection{Floorplan Localization}
FLoc tasks are often associated with LiDAR-based \textbf{M}onte \textbf{C}arlo \textbf{L}ocalization (MCL) \citep{dellaert1999monte, chu2015you, mendez2018sedar, winterhalter2015accurate}, which is a classical framework for 2D localization on purely geometric maps. However, the usage of LiDAR hinders the application of such localization algorithms on common mobile devices. To alleviate this limitation, some work \citep{boniardi2019robot, chu2015you, howard2022lalaloc++, howard2021lalaloc, min2022laser} investigate visual FLoc based on monocular and panoramic images. Some of these methods leverage 2D scene priors \citep{boniardi2019robot} and visual features \citep{min2022laser} by matching them with scene layouts to achieve visual FLoc. Several other methods \citep{howard2022lalaloc++, howard2021lalaloc} localize by comparing the panoramic image features rendered at specific locations with the query image features. However, these methods either assume known camera and room heights or require panoramic images, which limits the generalization of the localization algorithms. Recently, researchers have been working on generic monocular vision FLoc techniques \citep{karkus2018particle, chen2024f3loc} that employ Bayesian filters \citep{jonschkowski2016end, van2000unscented, bishop2001introduction} to solve the long sequence FLoc problem. Representatively, F$^3$Loc \citep{chen2024f3loc} decouples the visual FLoc task into a front-end observation model and a back-end histogram filter, achieving SoTA localization success and accuracy. 

In addition, Laser \citep{min2022laser} and SeDAR \citep{mendez2020sedar} utilize additional semantic information such as windows to assist visual FLoc. However, such semantic information is not always available in floorplans, thus our work only considers geometric occupancy information. Inspired by existing methods \citep{howard2021lalaloc, min2022laser, chen2024f3loc}, our work employs a front- and back-end FLoc framework that localizes by actively comparing the predicted pose features (i.e., 2D rays similar to LIDAR scans) with the floorplan. Unlike existing work \citep{boniardi2019robot, chu2015you, howard2022lalaloc++, howard2021lalaloc, min2022laser, chen2024f3loc} that leverages 2D geometric priors, our work incorporates 3D geometric priors into monocular visual FLoc for the first time, achieving significant performance gains.

\vspace{-0.3cm}
\subsection{Self-Supervised Scene Prior Modeling}
Self-supervised scene prior modeling can be classified into methods based on unimodal \citep{yadav2023offline, du2021curious, chen2020simple, chen2023think, zhu2024spa} and multimodal \citep{hong2023learning, arsomngern2023learning, chen2022self, chen2024embodied, zhang2024towards} contrastive learning. Unimodal schemes typically summarize 2D scene priors from extensive visual images \citep{yadav2023offline, du2021curious, chen2020simple, zhu2024spa} and employ diverse data enhancement techniques to construct positive/negative sample pairs. The difference is that offline methods \citep{yadav2023offline, chen2020simple, zhu2024spa} only utilize static and fixed datasets, whereas embodied methods \citep{du2021curious} can actively explore novel surroundings. To alleviate the view occlusion caused by 3D objects and obtain geometric priors beyond images, Chen et al. \citep{chen2023think} use offline contrastive learning to extract continuous relations between objects from 2D semantic maps. Unlike the above methods, one of our main insights is to bridge the modal gap between visual images and floorplans using 2D-3D crossmodal geometric priors.

Multimodal methods gain attraction due to their ability to share modality-specific contexts. For example, $Ego^2$-Map \citep{hong2023learning} proposes to learn scene priors by aligning egocentric views with 2D semantic maps in a crossmodal manner. 3DLFVG \citep{zhang2024towards} achieves visual grounding by aligning the 2D geometric relations in RGB images with the spatial relations between 3D objects in the point cloud. Several other 2D-3D crossmodal methods \citep{arsomngern2023learning, chen2022self, chen2024embodied} inject 3D geometric priors into the 2D visual models by semantically or spatially aligning RGB image features with the matched point cloud sets. However, the constraints used by these methods \citep{hong2023learning, arsomngern2023learning, chen2022self, chen2024embodied, zhang2024towards} to model 3D geometric priors are slack, i.e., the features of different modalities are roughly aligned at the set level. We believe that such soft constraints are not suitable for visual FLoc tasks that require high geometric accuracy. Therefore, we propose two 3D scene prior modeling techniques with hard geometric constraints and demonstrate their superiority using thorough comparative studies.

\begin{figure*}[t]
    \centering
    \includegraphics[scale=1.05]{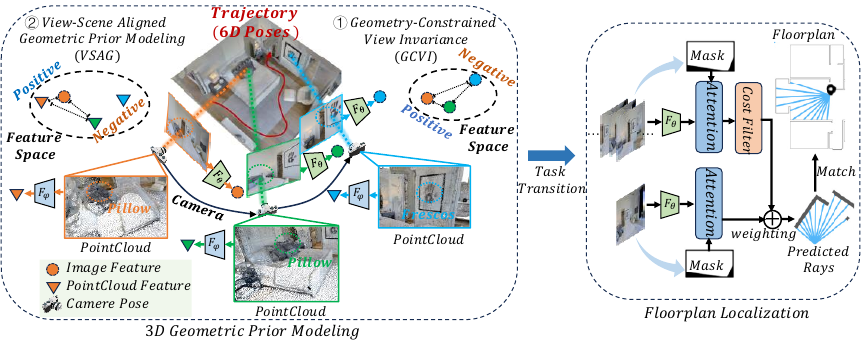}
    \vspace{-0.5cm}
    \caption{Our contrastive pre-training includes the modelings of GCVI ($\S$\ref{sub3.2}) and VSAG prior ($\S$\ref{sub3.3}). The GCVI is modeled by leveraging imaging geometric principles to provide matching constraints between multiple images that see the same points. The VSAG prior is modeled to enhance the crossmodal geometry-color correspondences by associating the scene's surface reconstruction with the RGB frames of the sequence. The pre-trained visual encoder $F_\theta$ is transferred to the downstream FLoc task for fine-tuning to further fit this task ($\S$\ref{sub3.4})}
    \label{fig2}
    \vspace{-0.5cm}
\end{figure*}

\vspace{-0.4cm}
\section{Methodology}

\subsection{Problem Definition and Overview} \label{sub3.1}

This work aims to cross-modally localize RGB images to specific imaging locations in a floorplan. The floorplan is a minimalist representation of a building's structure, which retains necessary geometric occupancy information such as doors and walls but no semantic categories. Given a time-varying image sequence $\mathcal{I}=\{\mathbf{I}_r|r \in \{ t-k, ..., t\}\}$ containing $k+1$ RGB images, we aim to find the current $SE(2)$ camera pose $s_t = [s_t^x, s_t^y, s_t^{\phi}]$ in a given 2D floorplan. It is worth noting that the camera intrinsics, the images' gravity directions, and the relative poses between the images are known. We investigate visual FLoc in two cases: \textbf{(1)} single-frame FLoc using only the current RGB observation $\mathbf{I}_t$, i.e., $k = 0$, and \textbf{(2)} using sequential multi-frame RGB images $\mathcal{I}$ to mitigate single-frame FLoc's uncertainty and ambiguity caused by repetitive structures and 3D objects. 

We propose to integrate 3D geometric priors into the FLoc policy to bridge the modal gap between visual images and floorplans while reducing localization biases caused by frequent visual changes and diverse indoor objects. Specifically, we introduce the modeling of GCVI and VSAG prior in Sec. \ref{sub3.2} and Sec. \ref{sub3.3}, respectively. Then, we present the 3D geometric priors enhanced visual FLoc method based on the F$^3$Loc framework in Sec. \ref{sub3.4}. The overview of our method is illustrated in Fig. \ref{fig2}.

\vspace{-0.4cm}
\subsection{Geometry-Constrained View Invariance} \label{sub3.2}
%从介绍细粒度的几何约束对于高精度定位的优势写起，其他的对比学习方法不具备这个优势
%利用多视图约束来学习视图不变性，而不需要昂贵的语义标签
Existing contrastive pre-trainings \citep{chen2020simple, he2020momentum, hong2023learning, arsomngern2023learning, chen2022self, chen2024embodied, zhang2024towards} usually use diverse data augmentations or crossmodal alignments to find positive matching pairs. Their core idea is to perform visual representation learning by bringing similar image/point cloud features in the feature space close to each other and moving different ones away from each other. Therefore, their feature embeddings are image/point cloud-aligned rather than pixel/point-aligned. Benefiting from massive image data \citep{chen2020simple, he2020momentum} and embodied learning paradigms \citep{du2021curious, liang2023alp}, these contrastive pre-trainings are powerful in representing scenes at the semantic level, facilitating a wide range of visual classification, detection, and segmentation tasks. 

However, in image sequences used for FLoc, the context between frames may changes little over time, which makes it hard to distinguish images into positive and negative sample pairs using data augmentations. In addition, we believe that the image and semantic-level constraints are too loose for FLoc as it geometrically requires high accuracy. Therefore, our GCVI is modeled by using the available 3D RGB-D scans and geometric knowledge to provide strict matching constraints between multiple images that see the same points. For this purpose, we employ the ScanNet dataset, which provides reconstructed scene surface geometry $S$ \citep{niessner2013real} and sequences of RGB-D images with camera poses. Notably, the poses and 3D reconstructions are obtained in a fully automated manner using an advanced SLAM method \citep{dai2017bundlefusion} without human intervention.

As shown in Fig. \ref{fig2}, our GCVI modeling utilizes 3D surface reconstruction to find pixel correspondences between 2D frame pairs. Specifically, for all depth image pairs $(\mathbf{D}_i, \mathbf{D}_j)$ in the RGB-D sequence, we use imaging geometric principles to project their depth values as 3D points. For example, given a 2D pixel coordinate $(u,v)$ in $\mathbf{D}_i$, we get the 3D point $P_w = [x,y,z]$ using a pinhole camera model $[u,v,\mathbf{D_i}(u, v)]=\mathbf{C}^{-1}([x,y,z])$:
% \vspace{-0.2cm}
{\setlength\abovedisplayskip{0.1cm}
\setlength\belowdisplayskip{0.1cm}
\begin{flalign} 
\begin{aligned}
\label{eq1}
P = \mathbf{K}^{-1} \cdot 
\left[
\begin{matrix}
u \cdot \mathbf{D_i}(u, v)  \\
v \cdot \mathbf{D_i}(u, v)  \\
\mathbf{D_i}(u, v)  \\
\end{matrix}
\right].
\end{aligned}
\end{flalign}
\begin{flalign} 
\begin{aligned}
\label{eq2}
P_w = \mathbf{R} \cdot P + \mathbf{T}.
\end{aligned}
\end{flalign}
where $\mathbf{K}$ denotes the camera intrinsics, $\mathbf{R}$ denotes the rotation matrix, and $\mathbf{T}$ denotes the translation vector. Then, the correspondence between two points from different frames is determined as their 3D coordinates lie within 2 cm of each other. This distance threshold is carefully chosen to balance accuracy and computational cost. We utilize pairs of frames with at least 30\% pixel correspondence, and the ratio is computed as the number of corresponding pixels in the two frames divided by the total number of points in the two frames. In total, we obtain about 840k image pairs from the ScanNet training data.

During training, as shown in Fig. \ref{fig2}, a pair of RGB images with the above correspondence is fed into a shared UNet-style \citep{ronneberger2015u} backbone $F_\theta$ with ResNet50 architecture \citep{he2016deep}. Their pixel-to-pixel correspondences mean that at least 30\% of their RGB pixels refer to the same 3D physical points. Notably, even though differences in lighting effects and viewing angles may result in different pixel colors in both frames, the hard geometric constraint ensures that these pixels represent the same physical coordinates. Such geometric properties are beneficial for high-accuracy FLoc. Therefore, we treat image pairs that have such correspondence as positive samples $(\mathbf{f}_i, \mathbf{f}_j)$ and those that do not as negative samples $(\mathbf{f}_i, \mathbf{f}_k)$ for GCVI contrastive learning. The PointInfoNCE loss \citep{xie2020pointcontrast} used to train $F_\theta$ is defined as:
 \begin{flalign} 
\begin{aligned}
\label{eq3}
&\ \mathcal{L}_{GCVI} = - \sum_{(i, j) \in M}log \dfrac{exp(\mathbf{f}_i \cdot \mathbf{f}_j/\tau)}{\sum_{(\cdot,k) \in M}exp(\mathbf{f}_i \cdot \mathbf{f}_k/\tau)}, &
\end{aligned}
\end{flalign}
where $M$ denotes the set of image pairs of pixel correspondences. $\mathbf{f}_{*}$ denotes the feature vector encoded by $F_\theta$. $\tau$ denotes the temperature coefficient. Notably, our GCVI modeling can leverage multi-view constraints to learn view invariants without the need for expensive semantic labels.

\vspace{-0.4cm}
\subsection{View-Scene Aligned Geometric Prior Modeling} \label{sub3.3}
%利用RGB-D数据固有的显式几何颜色对应关系，从RGB-D重建的几何表示中学习特征
In addition to the view-invariance constraint, we explicitly employ the inherent color-geometry correspondences between RGB images and 3D surface reconstructions to model the VSAG prior. Technically, each frame in the RGB-D image sequence is projected into world space based on the imaging geometric principles (Eq. (\ref{eq1}) and Eq. (\ref{eq2})) to compute its view frustum. Then, we crop the partial volumetric chunk $V_i = \{ \mathbf{C}(u, v, \mathbf{D_i}(u, v))\ |\ 0 \le u < w, 0 \le v < h \}$ that lies within the frustum from the 3D surface $S$ along the view frustum's axis-aligned orientation. As shown in Fig. \ref{fig2}, the volume chunk $V_i$ is represented as a point cloud with a resolution of 2 cm. Notably, the pixel-point associations are established by projecting the depth values in $D_i$ corresponding to RGB pixels into the world space. Such kind of association is also a hard constraint, since the projected points are matched to the corresponding points on the 3D surface within an error of 2 cm. A pair of color frame and geometric chunk $(\mathbf{I}_i, \mathbf{V}_i)$ with the above constraint is treated as a positive sample pair for contrastive learning to model the VSAG prior.

In practice, we employ a UNet-style backbone $F_\theta$ with ResNet50 architecture and PointNet++ \citep{qi2017pointnet++} $F_\varphi$ to model the color-geometry correspondence. It is worth noting that $F_\theta$ is shared with the view-invariant prior modeling. Similar to the GCVI contrastive learning, samples with the above color-geometry associations are treated as positive pairs and those without as negative pairs, as shown in Fig. \ref{fig2}. PointInfoNCE is likewise used as the contrastive loss $\mathcal{L}_{VSAG}$. The difference is that $\mathbf{f}_i$ denotes 2D image features, $\mathbf{f}_j$ denotes the corresponding 3D point cloud features, and $M$ denotes the set of 2D-3D pixel-point correspondence pairs. The modeling of VSAG prior relies on 2D-3D geometric constraints provided by RGB-D scans, which likewise require no manual annotations. The advantage of the VSAG prior modeling is that it fully leverages strict view-geometry alignments to enhance the FLoc algorithm's sensitivity to visual changes induced by 3D objects. Such visual changes include not only geometric changes but also color changes, as the 2D colored textures in the RGB frames are mapped to the 3D details in the RGB-D scans.

\vspace{-0.5cm}
\subsection{3D Priors Enhanced Visual FLoc} \label{sub3.4}
%介绍基于FLoc的视觉平面图定位的技术细节，
By modeling view invariance and view-scene aligned geometric priors, we implicitly and cross-modally inject 3D priors into the 2D visual encoder $F_\theta$. As shown in Fig. \ref{fig2}, the fully pre-trained visual encoder is transferred to the visual FLoc task for fine-tuning to fit the task. As described in Sec. \ref{sub3.1}, we investigate the single-frame and multi-frame visual FLoc techniques in the F$^3$Loc \citep{chen2024f3loc} framework, which consists of an observation model and a histogram filter \citep{jonschkowski2016end}. F$^3$Loc localizes by finding the pose in the floorplan that has the most similar 2D rays (similar to LIDAR scans) as the prediction, as shown in Fig. \ref{fig2}. We introduce our FLoc method by reviewing the front- and back-end architecture of F$^3$Loc:

\textit{\textbf{Observation Model.}} For both single-frame and multi-frame visual FLoc, F$^3$Loc first aligns the image with the gravity direction and uses $F_{\theta}$ and an attention \citep{vaswani2017attention} based network to learn the probability distribution of planar depth over a range of depth hypotheses. Pixels that become unobservable due to gravity alignment are masked in the attention. As shown in Fig. \ref{fig2}, the difference is that the multi-frame visual FLoc employs an additional 2D convolution-based learnable cost filter to compute the final floorplan depth from cross-view features. For both visual FLoc settings, an equiangular ray scan is constructed from the predicted floorplan depth to localize in the floorplan. To unify the two visual FLoc settings in a single framework, a multilayer perceptron is adopted to learn a weight $0 \le \omega \le 1$ from the two predictions for soft selection:
\begin{flalign} 
\begin{aligned}
\label{eq4}
\mathbf{P}_{fuse}=\omega Upsample(\mathbf{P}_{single}) + (1 - \omega)\mathbf{P}_{mv}.
\end{aligned}
\end{flalign}
$\mathbf{P}_{single}$ and $\mathbf{P}_{mv}$ denote the probability distributions of planar depth from a single frame and multiple frames, respectively. The upsampling operation is used to align the dimensions. The expectation of $\mathbf{P}_{fuse}$ provides the final prediction of 2D rays. $\omega$ is manually specified as 1 and 0 implying that only single-frame and multi-frame visual FLoc are used, respectively.

In our work, the 3D geometric priors learned through contrastive pre-training are injected into the observation model without increasing the computational overhead of the visual FLoc algorithm. The geometric constraints established by projecting RGB images into the 3D world space facilitate bridging the modal gap between visual observations and floorplans, which further boosts the success and accuracy of visual FLoc.

\textit{\textbf{Histogram Filter.}} Similar to F$^3$Loc, we also leverage a histogram filter \citep{jonschkowski2016end} to keep track of the localization posterior over the entire floorplan. Such a filtering scheme is particularly effective in dealing with long sequential FLoc, as shown in Fig. \ref{fig4} and Fig. \ref{fig5}. 

\textit{\textbf{Loss Function.}} For the training of FLoc models, we optimize an L1 loss and a cosine similarity-based shape loss:
\begin{flalign} 
\begin{aligned}
\label{eq5}
\mathcal{L}_{FLoc} = ||\mathbf{d}, \mathbf{d}^{*}||_1 + \frac{\mathbf{d}^{\top}\mathbf{d}^{*}}{max\{||\mathbf{d}||_2||\mathbf{d}^{*}||_2,\epsilon \}}.
\end{aligned}
\end{flalign}
Where $\mathbf{d}$ and $\mathbf{d}^{*}$ are predicted and \textbf{G}round \textbf{T}ruth (GT) 2D-ray depths, respectively. $\epsilon$ is a small constant to prevent division by zero.

\vspace{-0.2cm}
\section{Experiments}

\subsection{Experimental Setup}

\textit{\textbf{Datasets.}} We first use the challenging Structured3D (full) \citep{zheng2020structured3d} dataset to perform comparative studies between our single-frame FLoc method $Ours_s$ and the SoTA methods. Structured3D is a photorealistic dataset containing 3296 fully furnished indoor environments with in total 78,453 perspective images. Notably, we use monocular images rather than panoramic images, and the horizontal field of view of each image is 80$^\circ$. The resolution of the floorplan extracted from the Structured3D dataset is 0.02 m. For model training and evaluation, we use the official data splits. 

\renewcommand\arraystretch{0.85}
\begin{table}[!t] \small
\caption{Comparative studies of single-frame visual FLoc methods on the Structured3D (full) dataset.}
\vspace{-0.4cm}
\label{table1}
\begin{center}
\setlength{\tabcolsep}{0.4mm}{
\begin{tabular}{c | c c c c }
\bottomrule[1.3pt]
\multirow{2}{*}{\textbf{Method \tiny{(Venue)}}} &
\multicolumn{4}{c}{\textbf{Structured3D (full)}} \\
\cline{2-5}
& \textbf{SR@0.1m}  & \textbf{SR@0.5m}  & \textbf{SR@1m}  & \textbf{SR@1m30$^\circ$} \\
\hline
PF-net \citep{karkus2018particle} \tiny{(CoRL 2018)} & 0.2 & 1.3 & 3.2 & 0.9  \\
MCL \citep{dellaert1999monte} \tiny{(ICRA 1999)} & 1.3 & 5.2 & 7.8 & 6.4 \\
LASER \citep{min2022laser} \tiny{(CVPR 2022)} & 0.7 & 6.4 & 10.4 & 8.7 \\
\rowcolor{green!10} F$^{3}$Loc$_s$ \citep{chen2024f3loc} \tiny{(CVPR 2024)} & 1.5 & 14.6 & 22.4 & 21.3 \\
\hline
\rowcolor{green!10} Ours$_s$ & \textbf{5.6} \scriptsize{(+4.1)} & \textbf{27.4} \scriptsize{(+12.8)} & \textbf{55.5} \scriptsize{(+33.1)} & \textbf{24.0} \scriptsize{(+2.7)}\\
\bottomrule[1.3pt]
\end{tabular}}
\end{center}
\vspace{-0.7cm}
\end{table}

\begin{table*}[!t] \small
\caption{Comparative studies between our single-frame ($Ours_s$), multi-frame ($Ours_m$), and adaptive ($Ours_f$) visual FLoc methods with baselines on Gibson(f) and Gibson(g) datasets.}
\vspace{-0.5cm}
\label{table2}
\begin{center}
\setlength{\tabcolsep}{0.9mm}{
\begin{tabular}{c c c c c | c c c c }
\bottomrule[1.3pt]
\multirow{2}{*}{\textbf{Method \tiny{(Venue)}}} &
\multicolumn{4}{c}{\textbf{Gibson(f)}} & \multicolumn{4}{c}{\textbf{Gibson(g)}} \\
\cline{2-9}
& \textbf{SR@0.1m}  & \textbf{SR@0.5m}  & \textbf{SR@1m} & \textbf{SR@1m30$^\circ$}  & \textbf{SR@0.1m}  & \textbf{SR@0.5m}  & \textbf{SR@1m}  & \textbf{SR@1m30$^\circ$}\\
\hline
PF-net \citep{karkus2018particle} \tiny{(CoRL 2018)} & 0 & 2.0 & 6.9 & 1.2 & 1.0 & 1.9 & 5.6 & 1.9 \\
MCL \citep{dellaert1999monte} \tiny{(ICRA 1999)} & 1.6 & 4.9 & 12.1 & 8.2 & 2.3 & 6.2 & 9.7 & 7.3 \\
LASER \citep{min2022laser} \tiny{(CVPR 2022)} & 0.4 & 6.7 & 13.0 & 10.4 & 0.7 & 7.0 & 11.8 & 9.5 \\
\rowcolor{green!10} F$^{3}$Loc$_s$ \citep{chen2024f3loc} \tiny{(CVPR 2024)} & 4.7 & 28.6 & 36.6 & 35.1 & 4.3 & 26.7 & 33.7 & 32.3 \\
\rowcolor{blue!10} F$^{3}$Loc$_m$ \citep{chen2024f3loc} \tiny{(CVPR 2024)} & 13.2 & 40.9 & 45.2 & 43.7 & 9.3 & 27.0 & 31.0 & 29.2\\ 
\rowcolor{red!10} F$^{3}$Loc$_f$ \citep{chen2024f3loc}
\tiny{(CVPR 2024)} & 14.3 & 42.1 & 47.4 & 45.6 & 12.2 & 39.4 & 44.5 & 43.2 
\\
\hline
\rowcolor{green!10} Ours$_s$ & 5.3 \scriptsize{(+0.6)} & 33.2 \scriptsize{(+4.6)} & 39.8 \scriptsize{(+3.2)} & 38.4 \scriptsize{(+3.3)} & 9.4 \scriptsize{(+5.1)}  & 37.4 \scriptsize{(+10.7)}  & 43.1 \scriptsize{(+9.4)}  & 41.5 \scriptsize{(+9.2)}  \\
\rowcolor{blue!10} Ours$_m$ & 15.3 \scriptsize{(+2.1)} & 42.5 \scriptsize{(+1.6)} & 47.4 \scriptsize{(+2.2)} & 45.9 \scriptsize{(+2.2)} & 11.2 \scriptsize{(+1.9)} & 36.3 \scriptsize{(+9.3)} & 41.6 \scriptsize{(+10.6)} & 39.8 \scriptsize{(+10.6)} \\
\rowcolor{red!10} Ours$_f$ & \textbf{16.0} \scriptsize{(+0.7)} & \textbf{45.2} \scriptsize{(+3.1)} & \textbf{50.0} \scriptsize{(+2.6)} & \textbf{48.7} \scriptsize{(+3.1)} & \textbf{13.7} \scriptsize{(+1.5)} & \textbf{41.5} \scriptsize{(+2.1)} & \textbf{46.4} \scriptsize{(+1.9)} & \textbf{44.5} \scriptsize{(+1.3)} \\
\bottomrule[1.3pt]
\end{tabular}}
\end{center}
\vspace{-0.5cm}
\end{table*}

\begin{table}[!t] 
\caption{Comparative studies of long-sequence visual Floc methods on the Gibson(t) dataset. *-Gibson(f) and *-Gibson(g) denote training the visual Floc models using Gibson(f) and Gibson(g) datasets, respectively.}
\vspace{-0.5cm}
\label{table3}
\begin{center}
\setlength{\tabcolsep}{0.4mm}{
\begin{tabular}{c | c c c c }
\bottomrule[1.3pt]
\multirow{2}{*}{\textbf{Method \tiny{(Venue)}}} &
\multicolumn{3}{c}{\textbf{Gibson(t)}} \\
\cline{2-5}
& \textbf{SR@0.2m} & \textbf{SR@1m} & \textbf{RMSE(Succ)} & \textbf{RMSE(All)} \\
\hline
LASER \citep{min2022laser} \tiny{(CVPR 2022)} & - & 59.5 & 0.39 & 1.96 \\
\rowcolor{green!10} F$^{3}$Loc$_s$ \citep{chen2024f3loc} \tiny{(CVPR 2024)} & 35.1 & 89.2 & 0.18 & 0.88 \\
\rowcolor{red!10} F$^{3}$Loc$_f$ \citep{chen2024f3loc} \tiny{(CVPR 2024)} & 62.2 & 94.6 & 0.12 & 0.51 \\
\hline
\rowcolor{green!10} Ours$_s$-Gibson(f) & 54.1 \scriptsize{(+19.0)} & 89.2 \scriptsize{(+0.0)} & 0.16 \scriptsize{(-0.02)} & 0.75 \scriptsize{(-0.13)} \\
\rowcolor{green!10} Ours$_s$-Gibson(g) & 70.3 \scriptsize{(+35.2)} & 97.3 \scriptsize{(+8.1)} & 0.13 \scriptsize{(-0.05)} & 0.35 \scriptsize{(-0.53)} \\
% \rowcolor{gray!20} Ours$_m$ & 56.8 & 94.6 & 0.14 & 0.51 \\
\rowcolor{red!10} Ours$_f$-Gibson(g) & \textbf{74.6} \scriptsize{(+12.4)} & \textbf{98.1} \scriptsize{(+3.5)} & \textbf{0.11} \scriptsize{(-0.01)} & \textbf{0.31} \scriptsize{(-0.20)} \\
\bottomrule[1.3pt]
\end{tabular}}
\end{center}
\vspace{-0.7cm}
\end{table}

In addition, we employ a series of Gibson \citep{xia2018gibson} datasets (Gibson(g), Gibson(f), and Gibson(t)) collected by F$^{3}$Loc to fully evaluate our single-frame ($Ours_s$), multi-frame ($Ours_m$), and adaptive ($Ours_f$) visual FLoc methods. We follow the data split in F$^{3}$Loc, including 108 training scenes, 9 validation scenes, and 9 test scenes. The horizontal field of view of the images in the Gibson datasets is 108$^\circ$. The resolution of the floorplan extracted from the Gibson datasets is 0.1 m. Gibson(g) consists of general motions (including in-place steering motions) and includes 49,558 pieces of sequential views, each of which contains 4 image frames. Gibson(f) consists of only forward motions and includes 24,779 pieces of sequential views, each of which likewise contains 4 image frames. Therefore, Gibson(g) is intuitively more complex and harder than Gibson(f). Gibson(t) consists of 118 pieces of long-sequence views, each of which contains 280 $\sim$ 5152 image frames.

\textit{\textbf{Baselines.}} We compare our method with the following FLoc baselines, none of them employing semantic labels: \textit{\textbf{(1) PF-net \citep{karkus2018particle}}} proposes a particle filter specialized for visual FLoc. Its observation model aims to learn the similarity between an image and the corresponding map patch. \textit{\textbf{(2) MCL \citep{dellaert1999monte}}} is the most popular framework for 2D localization on pure geometry maps. In this work, we follow LASER \citep{min2022laser} to simulate a 72-ray 2D LiDAR giving ground-truth distance without noise as its input. \textit{\textbf{(3) LASER \citep{min2022laser}}} represents the floorplan as a set of points and gathers the features of the visible points of each pose in the floorplan. It actively compares the rendered pose features with the query image features for visual FLoc. \textit{\textbf{(4) F$^3$Loc \citep{chen2024f3loc}}} is the SoTA visual FLoc method that proposes a probabilistic model consisting of a ray-based observation module and a histogram filtering module. F$^3$Loc includes three variants: single-frame (F$^3$Loc$_s$), multi-frame (F$^3$Loc$_m$), and adaptive (F$^3$Loc$_f$) visual FLoc methods. 

\textit{\textbf{Evaluation Metrics.}} The \textbf{S}uccess \textbf{R}ates (SR) at different localization accuracies are used as evaluation metrics for comparative studies. For example, SR@0.1m implies the localization accuracy at a precision of 0.1 m. In addition, SR@1m30$^\circ$ denotes the success rate when the localization accuracy is 1 m and the orientation error is less than 30$^\circ$. For comparisons on the Gibson(t) dataset, the \textbf{R}oot-\textbf{M}ean-\textbf{S}quare \textbf{E}rror (RMSE) is also employed to measure the accuracy of sequential trajectory tracking when localization is successful (RMSE (Succ)) and in all cases (RMSE (All)).

\textit{\textbf{Implementation Details.}} For the pre-training of 3D geometric prior modeling, we use an Adam \citep{kingma2014adam} optimizer with a learning rate of 0.01 and a batch size of 4. The learning rate is decreased by a factor of 0.99 every 1000 steps, and our method is trained for 60,000 iterations. The constrastive losses $\mathcal{L}_{GCVI}$ and $\mathcal{L}_{VSAG}$ have a weighting of 1:1. The temperature coefficient $\tau$ is set to 0.07. 

For the fine-tuning of visual FLoc, we use an Adam optimizer with a learning rate of 0.001 for all training. For Structured3D, the single-frame FLoc model $Ours_s$ is trained for 100 epochs. For the Gibson dataset, the single-frame and multi-frame FLoc models ($Ours_s$ and $Ours_f$) are trained on the entire training split of Gibson(f) for 100 and 20 epochs, respectively. For the selection network, another pair of single-frame and multi-frame FLoc models are trained on 80 scenes of the training split before freezing their weights to train the selection network on the remaining 20 scenes of Gibson(g) for 5 epochs. The selection network is trained on disjoint scenes to prevent it from being biased by both modules’ performance on the visited scenes. All model training is performed on 4 NVIDIA 3090 GPUs. Our single-frame and multi-frame FLoc methods match the predicted 40 and 160 rays to the floorplans for localization, respectively. The cost filter in Fig. \ref{fig2} is implemented as a UNet-style \citep{ronneberger2015u} network that converts multi-channel image features into single-channel features for ray prediction. The selection network is implemented as 3 stacked linear layers with BatchNorm \citep{ioffe2015batch} and ReLU activation.
%还需要介绍用了多少条射线，以及平面图的分辨率，还有神经网络的细节（参考F3Loc的补充材料）

\vspace{-0.3cm}
\subsection{Comparative Studies}
We first compare the single-frame visual FLoc method $Ours_s$ with the existing SoTA methods on the Structured3D (full) dataset, and the experimental results are shown in Tab. \ref{table1}. Our method dramatically improves localization success rates at various precision levels. Quantitatively, our method improves SR@0.1m, SR@0.5m, SR@1m, and SR@1m30$^\circ$ by +4.1\%, +12.8\%, +33.1\%, and +2.7\% , respectively, relative to F$^3$Loc$_s$. Although PF-net, MCL, LASER, and F$^3$Loc$_s$ introduce 2D geometric cues into the observation modules, they typically cannot handle the complicated layouts in the large-scale Structured3D dataset. In contrast, our approach benefits from view invariance prior and view-scene alignment, which are modeled with full consideration of the presence of diverse furniture in the room. The significant performance gains imply that such 3D priors compensate for the modal and geometric differences between visual observations and floorplans.

In addition, we conduct comparative studies of the proposed methods against baselines on Gibson(f) and Gibson(g) datasets, and the experimental results are shown in Tab. \ref{table2}. Our approach significantly boosts the performance of single-frame $Ours_s$ and multi-frame $Ours_m$ FLoc methods on the Gibson(f) dataset. Since $Ours_f$ is trained to adaptively select single-frame and multi-frame visual FLoc methods based on visual observations, $Ours_f$ further benefits from the performance gains of $Ours_f$ and $Ours_m$. The experimental results in the first 6 rows of Tab. \ref{table2} show that the baseline methods perform better on the Gibson(f) dataset since Gibson(g) is more challenging than Gibson(f) (Gibson(f) contains only forward motions). Interestingly, the performance gains of $Ours_s$ on Gibson(g) are more significant compared to those on Gibson(f), even though Gibson(g) contains more complex in-place steering motions. Quantitatively, $Ours_s$ improves SR@0.1m, SR@0.5m, SR@1m, and SR@1m30$^\circ$ by +5.1\%, +10.7\%, +9.4\%, and +9.2\%, respectively, relative to F$^3$Loc$_s$ on Gibson(g). We owe these significant strengths to our 3D geometric prior modeling, i.e., geometry-constrained view invariance and view-scene aligned geometric prior. GCVI modeling enhances our method's robustness to drastic visual changes caused by in-place steering. VSAG prior modeling enhances our method's sensitivity to changes in visual appearance induced by 3D objects.

Although $Ours_m$ performs less well on the more challenging Gibson(g) than it does on Gibson(f), the improvement in its localization SR relative to F$^3$Loc$_m$ remains dramatic. Quantitatively, $Ours_m$ improves SR@0.1m, SR@0.5m, SR@1m, and SR@1m30$^\circ$ by +1.9\%, +9.3\%, +10.6\%, and +10.6\%, respectively, relative to F$^3$Loc$_m$ on Gibson(g). Furthermore, $Ours_f$ becomes stronger as a result of the enhancements of $Ours_s$ and $Ours_m$.

\begin{table}[t] \small
\caption{Ablation studies of 3D geometric priors.}
\vspace{-0.5cm}
\label{table4}
\begin{center}
\setlength{\tabcolsep}{1.5mm}{
\begin{tabular}{c c | c c c c}
\bottomrule[1.3pt]
\multicolumn{2}{c}{\textbf{Ablations}}&
\multicolumn{4}{c}{\textbf{Gibson(g)}}\\
\hline
GCVI & VSAG & \textbf{SR@0.1m}  & \textbf{SR@0.5m}  & \textbf{SR@1m}  & \textbf{SR@1m30$^\circ$} \\
\hline
\checkmark &  & 6.2 & 31.8 & 38.3 & 37.2 \\
& \checkmark & 6.4 & 32.2 & 37.5 & 36.0 \\
\rowcolor{gray!20} \checkmark & \checkmark & \textbf{9.4}  & \textbf{37.4}  & \textbf{43.1} & \textbf{41.5} \\
\bottomrule[1.3pt]
\end{tabular}}
\end{center}
\vspace{-0.5cm}
\end{table}

\begin{table}[!t] \small
\caption{Comparative studies of enhancing visual Floc by using different 2D/3D contrastive pre-trainings.}
\vspace{-0.5cm}
\label{table5}
\begin{center}
\setlength{\tabcolsep}{0.7mm}{
\begin{tabular}{c | c c c c }
\bottomrule[1.3pt]
\multirow{2}{*}{\textbf{Method \tiny{(Venue)}}} &
\multicolumn{4}{c}{\textbf{Gibson(g)}} \\
\cline{2-5}
& \textbf{SR@0.1m}  & \textbf{SR@0.5m}  & \textbf{SR@1m}  & \textbf{SR@1m30$^\circ$}\\
\hline
$Scratch$ & 4.3 & 21.8 & 27.9 & 26.2 \\
$IN$-$Pretraining$ & 4.3 & 26.7 & 33.7 & 32.3  \\
\hline
SimCLR \citep{chen2020simple} \tiny{(ICML 2020)} & 4.7 & 28.2 & 35.3 & 34.6 \\
CRL \citep{du2021curious} \tiny{(ICCV 2021)} & 5.0 & 29.7 & 37.2 & 35.8 \\
Ego$2$-MAP \citep{hong2023learning} \tiny{(ICCV 2023)} & 5.7 & 30.6 & 36.9 & 35.2 \\
ECL \citep{chen2024embodied} \tiny{(ACM MM 2024)} & 7.1 & 34.8 & 40.5 & 38.7 \\
SPA \citep{zhu2024spa} \tiny{(ICLR 2025)} & 8.3 & 35.7 & 41.4 & 39.5 \\
\hline
\rowcolor{gray!20} Ours & \textbf{9.4} \scriptsize{(+1.1)} & \textbf{37.4} \scriptsize{(+1.7)} & \textbf{43.1} \scriptsize{(+1.7)} & \textbf{41.5} \scriptsize{(+2.0)}  \\
\bottomrule[1.3pt]
\end{tabular}}
\end{center}
\vspace{-0.7cm}
\end{table}

We employ well-trained visual FLoc models to solve the long-sequence trajectory tracking problem on the Gibson(t) dataset, and the experimental results are shown in Tab. \ref{table3}. Technically, we combine the histogram filter proposed by F$^3$Loc with our 3D prior enhanced observation model. Although F$^3$Loc$_s$ trained with Gibson(g) achieves promising localization SR and accuracy, our model trained with Gibson(f) (only contains forward motions) performs better (line 2 vs. line 4). Such experimental results reflect that our 3D prior modelings provide more valuable geometric information, even employing training data of lower complexity. As expected, $Ours_s$ trained with Gibson(g) more significantly improves F$^3$Loc$_s$'s performance. Notably, $Ours_s$ dramatically improves the localization SR by +35.2\% at a precision of 0.2 m, reflecting our method's strength in localization accuracy. Similar to the performance on the Gibson(f) and Gibson(g) datasets, $Ours_f$-Gibson(g) achieves the best performance on the Gibson(t) dataset. Fig. \ref{fig3} illustrates the changes in success rate versus precision for long-sequence trajectory tracking using different numbers of historical frames. Our method outperforms F$^3$Loc in almost all localization precisions.

\begin{figure}[t]
    \centering
    \includegraphics[scale=0.6]{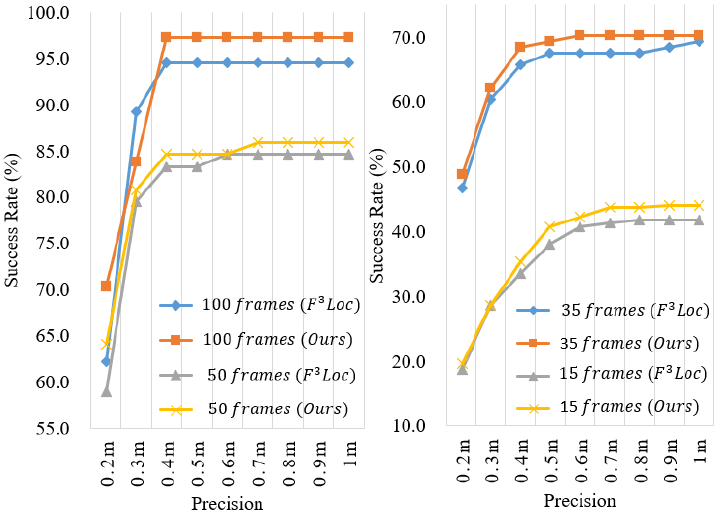}
    \vspace{-0.4cm}
    \caption{Comparison of the localization success rate using different number of historical frames. The more frames are used within the filter, the higher the localization success rate.}
    \label{fig3}
    \vspace{-0.6cm}
\end{figure}

\begin{figure}[t]
    \centering
    \includegraphics[scale=0.6]{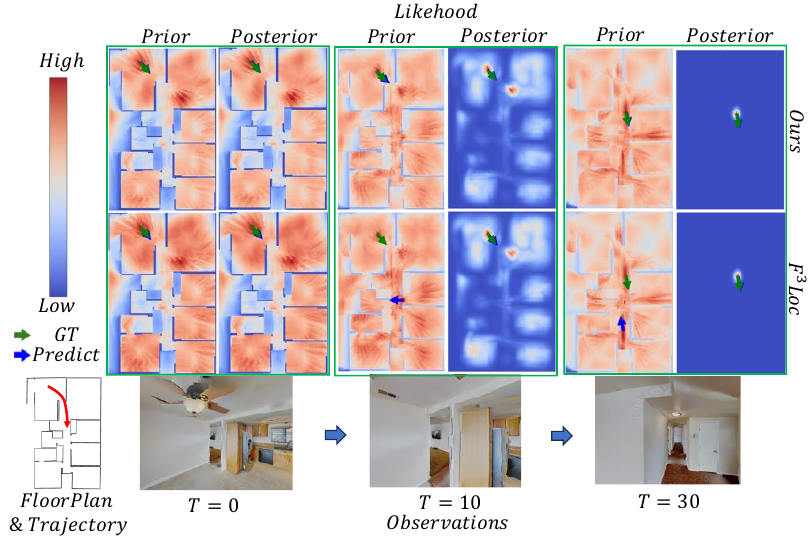}
    \vspace{-0.4cm}
    \caption{Qualitative comparative studies in a scene without complex furniture. $Prior$ illustrates the single-frame localization likelihood based on the current observation. $Posterior$ illustrates the likelihood of tracking a long sequence trajectory using a histogram filter.}
    \label{fig4}
    \vspace{-0.7cm}
\end{figure}

\begin{figure*}[t]
    \centering
    \includegraphics[scale=1.07]{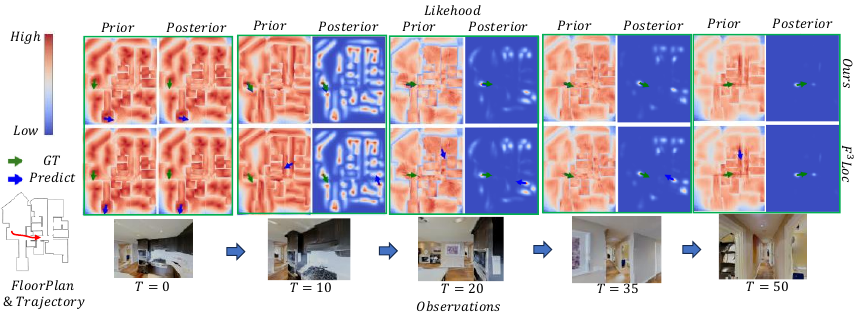}
    \vspace{-0.4cm}
    \caption{A qualitative comparative study in a larger-scale scene with complex furniture. Our single-frame localization is highly consistent with GT in most cases. Our long-sequence trajectory tracking quickly syncs the predicted pose to the same modality with GT, whereas F$^3$Loc shows long-term localization errors.}
    \label{fig5}
    \vspace{-0.5cm}
\end{figure*}

\vspace{-0.2cm}
\subsection{Ablation studies}
We conduct ablation studies on the modeling of GCVI and VSAG prior by employing the challenging Gibson(g) dataset and our single-frame visual FLoc method $Ours_s$. The experimental results are shown in Tab. \ref{table4}. We find that the absence of both 3D geometric priors can cause significant decreases in FLoc performance. Ablation studies show that the contributions of geometry-constrained view invariance and view-scene aligned geometric prior are almost equal, i.e., the localization success rates are roughly equal. However, the VSAG prior can lead to higher localization accuracy, i.e., higher SR@0.1m and SR@0.5m metrics.

In addition, we replace our 3D prior modelings with existing 2D/3D contrastive pre-trainings and neural rendering based embodied representation to highlight our approach's superiority. The experimental results are shown in Tab. \ref{table5}. $Scratch$ indicates that ResNet50 is trained for visual FLoc from scratch without using any geometric priors. $IN$-$Pretraining$ means using the ResNet50 pre-trained on ImageNet for visual FLoc. SimCLR, CRL, Ego$^2$-MAP, and ECL are all self-supervised scene representation learning methods based on the contrastive learning technique. The difference is that SimCLR's contrastive pre-training is limited to general images without introducing room-related geometric priors. CRL is the first proposal to learn visual representations in indoor scenes through embodied contrastive learning. Ego$^2$-MAP and ECL propose to align visual images with 2D/3D semantic maps to model 2D/3D geometric and semantic priors. Although ECL's 3D priors modeling is carried out in an embodied manner, it does not explicitly incorporate strict geometry constraints and only performs a coarse view-scene alignment. As a paradigm for neural rendering, SPA internalizes semantic and depth information into the visual encoder using advanced multi-view based volumetric rendering.

The experimental results in Tab. \ref{table5} show that different 2D/3D scene priors can improve visual FLoc's SR and accuracy to different degrees. Among them, SPA achieves significant performance gains by leveraging differentiable neural rendering on multi-view images. However, our approach achieves the best SR and accuracy in visual FLoc by making full use of 3D geometric priors. Quantitatively, our method improves SR@0.1m, SR@0.5m, SR@1m, and SR@1m30$^\circ$ by +1.1\%, +1.7\%, +1.7\%, and +2.0\%, respectively, relative to the strong baseline SPA. Such experimental results reflect the superiority of our contrastive pre-training based on hard geometric constraints, which is more applicable to visual FLoc tasks.

\vspace{-0.3cm}
\subsection{Qualitative Analysis}
We qualitatively compare our method with the SoTA method F$^3$Loc in challenging and simple scenes, respectively. As shown in Fig. \ref{fig4}, the simple scene does not contain complex furniture. Our method is more accurate in single-frame visual Floc and is on par with F$^3$Loc in long-sequence trajectory tracking. Since our approach explicitly models the GCVI, it is more robust to visual changes. Our VSAG prior accurately matches scene surface details to visual textures, which improves single-frame FLoc accuracy. Since it is relatively easy to cross-modally localize from a visual observation to a floorplan without the interference of 3D objects, both perform well in long-sequence visual FLoc. 

Fig. \ref{fig5} illustrates a qualitative comparison in the presence of complex furniture. In this case, our method performs significantly better in both single-frame visual FLoc and long-sequence trajectory tracking. In particular, our single-frame localization is highly consistent with GT in most cases. In addition, our long-sequence trajectory tracking quickly syncs the predicted pose to the same modality with GT, whereas F$^3$Loc shows long-term localization errors. We attribute these advantages to the modeling of 3D geometric priors. On the one hand, GCVI makes our method more robust to time-varying visual changes caused by complex furniture. On the other hand, the VSAG prior makes our method more sensitive to visual appearance changes induced by variously shaped 3D objects.

\vspace{-0.3cm}
\section{Conclusion and Discussion}
This paper focuses on the visual FLoc task and proposes two novel 3D geometric prior modeling techniques to enhance FLoc algorithms' observation model. Unlike existing methods that utilize 2D geometric priors and visual features, our method leverages 3D scene priors to bridge the modal gap between RGB images and floorplans. By integrating 3D geometric priors into the observation model, the success and accuracy of visual FLoc are significantly improved without increasing computational overhead. Our 3D geometric prior modeling uses hard constraints to construct positive/negative sample pairs for contrastive learning without the need of expensive labels and annotations. In addition, our method explicitly addresses the view occlusions and visual changes caused by 3D objects, which are not discussed in existing visual FLoc methods. Sufficient comparative studies reflect that our method achieves SoTA visual FLoc performance on Structured3D (full) and Gibson datasets. Ablation studies and qualitative analyses further demonstrate the potential of our 3D geometric prior modeling on visual FLoc. 

Despite achieving significant performance gains, the lack of indoor datasets with floorplans hinders the progress of FLoc. We believe that the domain gap can be mitigated and the generalization can be enhanced by internalizing 3D geometric priors from extensive realistic RGB-D scans into visual FLoc algorithms. While our method effectively uses 3D geometric cues, utilizing semantic information from images and floorplans can further reduce localization ambiguity. We will validate this idea in future work.

\begin{acks}
This work was supported in part by the National Natural Science Foundation of China under 62272489, 62332020, and 62350004, in part by the Natural Resources Science and Technology Plan Project of Hunan Province under 2021-17, and in part by the Open Competition Project of Xiangjiang Laboratory under 23XJ01011. This work was carried out in part using computing resources at the High-Performance Computing Center of Central South University.
\end{acks}

\bibliographystyle{ACM-Reference-Format}
\balance
\bibliography{cites}

\end{document}